\documentclass[twoside,11pt]{article}

\usepackage{jmlr2e}
\usepackage{listings}
\usepackage[]{mcode}

\jmlrheading{1}{2015}{1-5}{4/00}{10/00}{Reshad Hosseini Mohamadreza Mash'al}

\ShortHeadings{MixEst Toolbox for Mixture Models}{Hosseini and Mash'al}
\firstpageno{1}

\begin{document}

\title{MixEst: An Estimation Toolbox for Mixture Models}

\author{\name Reshad Hosseini \email reshad.hosseini@ut.ac.ir \\
	\addr School of ECE, College of Engineering\\
	University of Tehran\\
	Tehran, Iran
	\AND
	\name Mohamadreza Mash'al \email mrmashal@ut.ac.ir \\
	\addr School of ECE, College of Engineering\\
	University of Tehran\\
	Tehran, Iran}

\editor{}

\maketitle


\begin{abstract}

Mixture models are powerful statistical models used in many applications ranging from density estimation to clustering and classification.
When dealing with mixture models, there are many issues that the experimenter should be aware of and needs to solve.
The MixEst toolbox is a powerful and user-friendly package for MATLAB that implements several state-of-the-art approaches to address these problems.
Additionally, MixEst gives the possibility of using manifold optimization for fitting the density model, a feature specific to this toolbox.
MixEst simplifies using and integration of mixture models in statistical models and applications.
For developing mixture models of new densities, the user just needs to provide a few functions for that statistical distribution and the toolbox takes care of all the issues regarding mixture models.
MixEst is available at \textit{visionlab.ut.ac.ir/mixest} and is fully documented and is licensed under GPL.
\end{abstract}


\section{Introduction}

Mixture models are an integrated and fundamental component in many machine learning problems ranging from clustering to regression and classification~\citep{McLPee00}.
Estimating the parameters of mixture models is a challenging task due to the need to solve the following issues in mixture modeling:
\begin{itemize}
\item Unboundedness of the likelihood: This problem occurs when one component gets a small number of data points and  its likelihood becomes infinite~\citep{ciuperca_penalized_2003}. 
\item Local maxima: The log-likelihood objective function for estimating the parameters of mixture models is non-concave and has many local maxima~\citep{ueda_split_2000}.
\item Correct number of components: In many applications, it is needed to find the correct number of components~\citep{khalili_variable_2007}.
\end{itemize}
Addressing these issues for a mixture density when it is not available in common mixture modeling toolboxes will cost a lot of time and effort for the experimenter.
MixEst addresses all these issues not only for already implemented densities, but also for densities that the user may implement.
By implementing densities, we mean implementing a few simple functions which will be briefly discussed in section \ref{sec.model}.

This toolbox provides a framework for applying manifold optimization for estimating the parameters of mixture models.
This is an important feature of this toolbox, because recent empirical evidence shows that manifold optimization can surpass expectation maximization in the case of mixtures of Gaussians \citep{hosseini2015}.
It also opens the door for large-scale optimization by using stochastic optimization methods.
Stochastic optimization also allows solving the likelihood unboundedness problem mentioned above, without the need of implementing a penalizing function for the parameters of the density.

While several libraries are available for working with mixture models, to the best of our knowledge, none of them offers a modular and flexible framework that allows for fine-tuning the model structure or can provide universal algorithms for estimating model parameters solving all the problems listed above.
A review of features available in some libraries can be seen in Section~\ref{sec.comparison}.

In the next section, we give a short overview of the toolbox and its features.

\section{About the MixEst Toolbox}
This toolbox offers methods for constructing and estimating mixtures for joint density and conditional density modeling, therefore it is applicable to a wide variety of applications like clustering, regression and classification through probabilistic model-based approach.
Each distribution in this toolbox is a structure containing a manifold structure representing parameter space of the distribution along with several function handles implementing density-specific functions like log-likelihood, sampling, etc.
Distribution structures are constructed by calling factory functions with some appropriate input arguments defining the distribution.
For example for constructing a mixture of one-dimensional Gaussians with 2 components, it will suffice to write the following commands in MATLAB:
\vspace{-0.5cm}
\begin{lstlisting}
Dmvn = mvnfactory(1);
Dmix = mixturefactory(Dmvn, 2);
\end{lstlisting}
As an example of how to evoke a function handle, consider generating 1000 samples from the previously defined mixture:
\vspace{-0.5cm}
\begin{lstlisting}
theta.D{1}.mu = 0; theta.D{1}.sigma = 1; % mean and variance of the 1st component
theta.D{2}.mu = 5; theta.D{2}.sigma = 2; % mean and variance of the 2nd component
theta.p = [0.8 0.2]; % weighting coefficients of components
data = Dmix.sample(theta, 1000);
\end{lstlisting}
Each distribution structure exposes a common interface that optimization algorithms in the toolbox can use to estimate its parameters.
In addition to the EM algorithm which is a commonly implemented method in available libraries, our toolbox also makes optimization on manifolds available featuring procedures like early-stopping and mini-batching to avoid overfitting.
For optimization on manifolds, our toolbox depends on optimization procedures of an excellent toolbox called Manopt \citep{boumal_manopt_2014}.
In addition to optimization algorithms of Manopt like steepest descent, conjugate gradient and trust regions methods, the user can also use our implementation of Riemmanian LBFGS method.

\section{Model Development}
\label{sec.model}
MixEst includes many joint and conditional distributions to model data ranging from continuous to discrete and also directional.
Some users, however, may want to apply the tools developed in this toolbox for mixtures of a distribution not available in the toolbox yet.
To this end, the user needs to write a factory function that constructs a structure for the new distribution.

Each distribution structure has a field named ``M" determining the manifold of its parameter space.
For example for the case of multivariate Gaussian distribution, this is a product manifold of a positive definite manifold and a Euclidean manifold:
\vspace{-0.5cm}
\begin{lstlisting}
% datadim is the function input argument determining the dimensionality of data
muM = euclideanfactory(datadim);
sigmaM = spdfactory(datadim);
D.M = productmanifold(struct('mu', muM, 'sigma', sigmaM));
\end{lstlisting}
The manifold of parameter space completely determines how parameter structure is given to or is returned by different functions.
The structure of parameters for multivariate Gaussian would have two fields, a  mean vector ``mu" and a covariance matrix ``sigma". 

To use the estimation tools of the toolbox, two main functions have to be implemented.
The \textit{weighted log-likelihood} (wll) function and a function for computing the gradient of sum-wll with respect to the distribution parameters.
The syntax for calling the wll function is:
\vspace{-0.5cm}
\begin{lstlisting}
llvec = D.llvec(theta, data);
\end{lstlisting}
The input argument \mcode{theta} is a structure containing the input parameters of the corresponding distribution.
The second input argument \mcode{data} can be either a data matrix or a structure having several fields such as the data matrix and weights, which is interpreted using the \mcode{mxe_readdata} function.
The output argument \mcode{llvec} is a vector with entries equal to wll for each datum (each column) in the data matrix.

The function to compute the gradient of sum-wll has the following syntax:
\vspace{-0.5cm}
\begin{lstlisting}
llgrad = D.llgrad(theta, data);
\end{lstlisting}
The input arguments are similar to the function \mcode{llvec}.
The output argument \mcode{llgrad} is a structure similar to the input argument \mcode{theta} returning the gradient of sum-wll with respect to each parameter.


Some other (optional) functions that can be implemented for distributions are:
\begin{itemize}

\item[] \mcode{init}: This is for initializing the estimator using the data. 

\item[] \mcode{estimatedefault}: If the maximum wll has a structure that allows fast optimization (or has a closed-form solution), this estimator can be implemented in this function.
When this function is not present, the Riemmanian optimization is called in the maximization step of EM algorithm.

\item[] \mcode{llgraddata}: This function computes the gradient of wll with respect to the data.
It is required in some special cases such as when the distribution is used as the radial component of an elliptically-contoured distribution or as the components in independent component analysis.

\item[] \mcode{ll}: This function is sum-wll (sum of the output vector of \mcode{llvec} function).
Sometimes it is faster to write this function differently than just calling \mcode{llvec} and summing up its output vector.

\end{itemize}
Two other functions that can be used in the split-and-merge algorithms to avoid local maxima of mixture models are \mcode{kl} (for computing KL-divergence) and \mcode{entropy} (for computing entropy).
If the user wants to evoke a maximum-a-posteriori estimate, the functions \mcode{penalizerparam}, \mcode{penalizercost} and \mcode{penalizergrad} need to be implemented. 

\section{Feature Comparison}
To demonstrate the richness of features in MixEst, we are comparing its features with several other well-known packages in Table~\ref{table:final-ll}.
Among many toolboxes available for mixture modeling, we select those that are feature-rich and representative. These packages are Sklearn~\citep{scikit-learn}, Mclust~\citep{fraley1999mclust}, FlexMix~\citep{Leisch_2004b}, Bayes Net~\citep{Murphy01thebayes} and MixMod~\citep{biernacki2006model}.
We include Bayes Net to demonstrate what a generic Bayesian graphical modeling toolbox can do. Sklearn is a powerful machine learning toolbox containing many tools, among others tools specific for mixture modeling. MixMod also provides bindings for Scilab and Matlab.



\begin{table}[htbp]
\caption{Feature comparison of our toolbox and some other well-known packages.
	Different rows correspond to the following specifications of different toolboxes:
	1. Programming language;
	2. Approaches for solving local minima problem (SM stands for split-and-merge approach, IDMM for infinite dirichlet mixture models, HC stands for initialization using hierarchical clustering);
	3. Manifold optimization;
	4. Bayesian approaches for inference (MAP stands for maximum-a-posteriori, VB stands for variational Bayes);
	5. Large-scale optimization (SEM stands for stochastic EM, MB stands for mini-batching);
	6. Having tools for model selection;
	7. Automatic model selection (CSM stands for competitive split-and-merge);
	8. Ease of extensibility;
	9. Having mixtures of experts;
        10. Having mixtures of classifiers;
	11. Having mixtures of regressors;   }
\label{table:final-ll}
\vskip 0.15in
\begin{center}
\begin{small}
\begin{sc}
\begin{tabular}{| l |c | c | c | c | c | c | }
\hline
~       & MixEst  & SKlearn  & Mclust  &  FlexMix & Bayes Net & MixMod\\ \hline
\# 1 & Matlab  & Python & R & R & Matlab & C++ \\ \hline
\# 2  & SM  & IDMM & HC& ---& ---& --- \\ \hline
\# 3 & Yes  & No & No &No & No & No\\ \hline
\# 4  & MAP  & VB & MAP & ---& MAP& SM\\ \hline
\# 5 & MB  & ---& --- &--- & ---& SEM\\ \hline
\# 6 & Yes  & No & Yes & No& No& Yes\\ \hline
\# 7 & CSM  & IDMM  & ---&--- & --- & ---\\ \hline
\# 8 & Easy  & --- & --- & Easy& Medium& ---\\ \hline
\# 9 & Yes  & No &No & No& Yes& No\\ \hline
\# 10 & Yes  & No & No& No & Yes& No\\ \hline
\# 11 & Yes  & No & No& Yes& Yes& No\\ \hline
\end{tabular}
\end{sc}
\end{small}
\end{center}
\vskip -0.1in
\end{table}
\label{sec.comparison}


\vskip 0.2in
\bibliography{mixest}

\begin{thebibliography}{11}
\providecommand{\natexlab}[1]{#1}
\providecommand{\url}[1]{\texttt{#1}}
\expandafter\ifx\csname urlstyle\endcsname\relax
  \providecommand{\doi}[1]{doi: #1}\else
  \providecommand{\doi}{doi: \begingroup \urlstyle{rm}\Url}\fi

\bibitem[Biernacki et~al.(2006)Biernacki, Celeux, Govaert, and
  Langrognet]{biernacki2006model}
Christophe Biernacki, Gilles Celeux, G{\'e}rard Govaert, and Florent
  Langrognet.
\newblock Model-based cluster and discriminant analysis with the mixmod
  software.
\newblock \emph{Computational Statistics and Data Analysis}, 51\penalty0
  (2):\penalty0 587--600, 2006.

\bibitem[Boumal et~al.(2014)Boumal, Mishra, Absil, and
  Sepulchre]{boumal_manopt_2014}
Nicolas Boumal, Bamdev Mishra, P.-A. Absil, and Rodolphe Sepulchre.
\newblock Manopt, a matlab toolbox for optimization on manifolds.
\newblock \emph{Journal of Machine Learning Research}, 15:\penalty0 1455--1459,
  2014.

\bibitem[Ciuperca et~al.(2003)Ciuperca, Ridolfi, and
  Idier]{ciuperca_penalized_2003}
Gabriela Ciuperca, Andrea Ridolfi, and J{\'e}r{\^o}me Idier.
\newblock Penalized maximum likelihood estimator for normal mixtures.
\newblock \emph{Scandinavian Journal of Statistics}, 30\penalty0 (1):\penalty0
  45--59, March 2003.

\bibitem[Fraley and Raftery(1999)]{fraley1999mclust}
Chris Fraley and Adrian~E Raftery.
\newblock Mclust: Software for model-based cluster analysis.
\newblock \emph{Journal of Classification}, 16\penalty0 (2):\penalty0 297--306,
  1999.

\bibitem[Hosseini and Sra(2015)]{hosseini2015}
Reshad Hosseini and Suvrit Sra.
\newblock Manifold optimization for {Gaussian} mixture models.
\newblock \emph{arXiv preprint arXiv:1506.07677}, 06 2015.

\bibitem[Khalili and Chen(2007)]{khalili_variable_2007}
Abbas Khalili and Jiahua Chen.
\newblock Variable selection in finite mixture of regression models.
\newblock \emph{Journal of the American Statistical Association}, 102\penalty0
  (479):\penalty0 1025--1038, September 2007.

\bibitem[Leisch(2004)]{Leisch_2004b}
Friedrich Leisch.
\newblock {FlexMix}: A general framework for finite mixture models and latent
  class regression in {R}.
\newblock \emph{Journal of Statistical Software}, 11\penalty0 (8):\penalty0
  1--18, 2004.

\bibitem[McLachlan and Peel(2000)]{McLPee00}
Geoffrey McLachlan and David Peel.
\newblock \emph{Finite mixture models}.
\newblock John Wiley and Sons, New Jersey, 2000.

\bibitem[Murphy(2001)]{Murphy01thebayes}
Kevin~P. Murphy.
\newblock The {Bayes Net} toolbox for matlab.
\newblock \emph{Computing Science and Statistics}, 33:\penalty0 2001, 2001.

\bibitem[Pedregosa et~al.(2011)Pedregosa, Varoquaux, Gramfort, Michel, Thirion,
  Grisel, Blondel, Prettenhofer, Weiss, Dubourg, Vanderplas, Passos,
  Cournapeau, Brucher, Perrot, and Duchesnay]{scikit-learn}
F.~Pedregosa, G.~Varoquaux, A.~Gramfort, V.~Michel, B.~Thirion, O.~Grisel,
  M.~Blondel, P.~Prettenhofer, R.~Weiss, V.~Dubourg, J.~Vanderplas, A.~Passos,
  D.~Cournapeau, M.~Brucher, M.~Perrot, and E.~Duchesnay.
\newblock Scikit-learn: Machine learning in python.
\newblock \emph{Journal of Machine Learning Research}, 12:\penalty0 2825--2830,
  2011.

\bibitem[Ueda et~al.(2000)Ueda, Nakano, Ghahramani, and
  Hinton]{ueda_split_2000}
Naonori Ueda, Ryohei Nakano, Zoubin Ghahramani, and Geoffrey~E. Hinton.
\newblock Split and merge {EM} algorithm for improving {Gaussian} mixture
  density estimates.
\newblock \emph{The Journal of VLSI Signal Processing}, 26\penalty0
  (1):\penalty0 133--140, 2000.

\end{thebibliography}

\end{document}